\documentclass[10pt,letterpaper]{article}
\usepackage[utf8]{inputenc}

\usepackage{cogsci}

 \cogscifinalcopy 

\usepackage{cmap}
\usepackage[T1]{fontenc}
\usepackage[american]{babel}
\usepackage{csquotes}
\usepackage{newtxtext,newtxmath}  

\usepackage[
  backend=biber,
  style=apa,
  natbib=true,
  annotation=false,
]{biblatex}
\usepackage{float} 

\usepackage[hidelinks]{hyperref}
\usepackage{CJKutf8}
\usepackage{booktabs}  
\usepackage{graphicx}  
\usepackage{multirow}  

\title{\textit{Hán Dān Xué Bù} (Mimicry) or \textit{Qīng Chū Yú Lán} (Mastery)? \\ A Cognitive Perspective on Reasoning Distillation in Large Language Models}

\author{
  {\fontsize{11}{13}\selectfont\bfseries Yueqing Hu$^{1,2}$, \quad Xinyang Peng$^{3}$, \quad Shuting Peng$^{4}$,} \\
  {\fontsize{11}{13}\selectfont\bfseries Hanqi Wang$^{5}$, \quad Tianhong Wang (wangtianhong@ahu.edu.cn)$^{1\dagger}$} \\
  
  \vspace{0.5em}
  
  {\normalsize 
   $^1$School of Philosophy, Anhui University, Hefei 230039, China \\
   $^2$Institute of Neuroscience, Chinese Academy of Sciences, Shanghai 200031, China \\
   $^3$Faculty of Education, University of Cambridge, Cambridge, UK \\
   $^4$School of Foreign Studies, South China Normal University, Guangzhou, China \\
   $^5$Division of Psychology and Language Sciences, University College London, London, UK
  } \\
  
  \vspace{0.5em}
  
  {\small $^{\dagger}$ corresponding author}
}

\begin{document}

\maketitle

\begin{abstract}
Recent Large Reasoning Models trained via reinforcement learning exhibit a ``natural'' alignment with human cognitive costs. However, we show that reasoning distillation via Supervised Fine-Tuning (SFT) fails to transmit this cognitive logic, leading to a ``Cargo Cult'' where students only mimic length. Testing the \textit{Hán Dān Xué Bù} (Superficial Mimicry) hypothesis across 14~models, we identify a ``Functional Alignment Collapse'': while teacher models mirror human difficulty scaling ($\bar{r}=0.64$), distilled students significantly degrade this alignment ($\bar{r}=0.34$). Crucially, they exhibit ``Negative Transfer,'' dropping below their own pre-distillation baselines. Our analysis reveals a ``Linear Inflation Law'' where students apply a constant verbosity multiplier ($\approx2.44$) regardless of complexity. Consequently, distillation decouples computational cost from cognitive demand, revealing that human-like cognition is an emergent property of active reinforcement, not passive imitation.

\end{abstract}

\section{Introduction}

A central goal of cognitive science and artificial intelligence is to construct systems that not only solve problems but do so via mechanisms analogous to human cognition. Recently, \citet{devarda2025cost} reported that Large Reasoning Models (LRMs) trained via Reinforcement Learning with Verifiable Rewards (RLVR) exhibit a ``natural'' alignment with human cognitive costs: the length of their Chain-of-Thought (CoT) reasoning traces predicts human reaction times across diverse domains, suggesting that goal-directed optimization can implicitly recover core features of human problem-solving complexity \citep{anderson1990adaptive}.

Yet the alignment reported by \citet{devarda2025cost} emerged from models that \textit{learned to reason} through autonomous exploration. In contrast, the prevailing paradigm for deploying efficient AI is \textbf{knowledge distillation} \citep{hinton2015distilling}, where smaller ``student'' models are trained via Supervised Fine-Tuning (SFT) to mimic the output traces of powerful ``teachers'' \citep{ho2023large}. This shift from \textit{learning by doing} to \textit{learning by imitating} raises a fundamental question: when a student learns to generate reasoning traces by copying a teacher, does it internalize the teacher's \textit{cognitive structure}, or merely mimic its \textit{linguistic style}?

Following \citet{brady2025dual}, we adopt dual-process theory \citep{kahneman2011thinking} as an \textit{organizing structure}---not as a claim about shared cognitive architecture---and frame two competing hypotheses:

\begin{enumerate}
    \item \textbf{The \textit{Hán Dān Xué Bù} (Superficial Mimicry) Hypothesis}.\footnote{\textit{Hán dān xué bù} (\begin{CJK*}{UTF8}{gbsn}邯郸学步\end{CJK*}) refers to someone who blindly imitates the walking style of people in Handan, only to fail and forget their own. It implies that student models might degrade their base capabilities while imperfectly mimicking the teacher.} Distillation induces a ``surface approach'' to learning \citep{marton1976qualitative}: the student acquires the \textit{form} of reasoning (verbosity) without the \textit{function} (dynamic resource allocation).
    
    \item \textbf{The \textit{Qīng Chū Yú Lán} (Functional Mastery) Hypothesis}.\footnote{\textit{Qīng chū yú lán} (\begin{CJK*}{UTF8}{gbsn}青出于蓝\end{CJK*}) means ``indigo blue is extracted from the indigo plant but is bluer than it,'' describing a student who surpasses their teacher.} Alternatively, the student may successfully compress the teacher's reasoning policy, maintaining or improving alignment with human cognitive costs relative to its size.
\end{enumerate}

We test these hypotheses by benchmarking 14 models---the DeepSeek-R1 teacher \citep{guo2025deepseek}, six distilled variants (Qwen/Llama backbones), and their Instruct-tuned counterparts---against human reaction times across six reasoning tasks adapted from \citet{devarda2025cost}. Our results support the \textit{Mimicry} hypothesis and identify a ``Linear Inflation Law'' under which students apply a fixed verbosity multiplier to their base outputs, decoupling computational cost from cognitive demand. Alignment with human cognition thus appears to be an emergent property of reinforcement learning, lost under the superficial imitation of supervised distillation.







\section{Methods}

\subsection{Datasets}

\begin{figure}[t]
  \centering
  \includegraphics[width=\columnwidth]{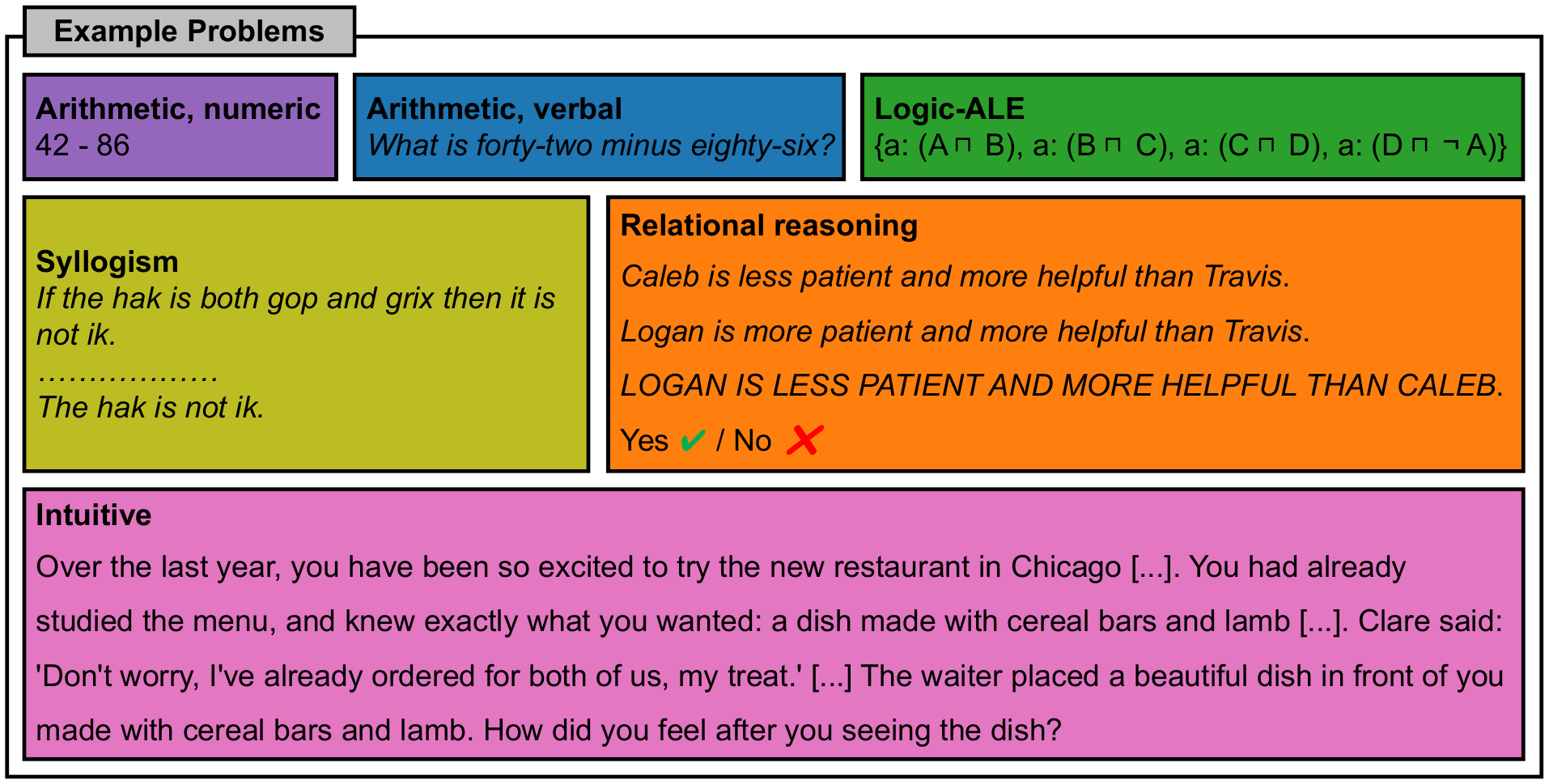} 
  \caption{Example reasoning problems.}
  \label{fig:task_demo}
\end{figure}

We utilized six reasoning tasks adapted from \citet{devarda2025cost}, spanning diverse cognitive domains (see Figure~\ref{fig:task_demo}). To verify alignment across a broad spectrum of human intelligence, we selected tasks ranging from algorithmic processing to common-sense intuition.\footnote{The \textit{H-ARC} task from the original study was excluded to strictly isolate linguistic reasoning costs. Recent scholarship argues that ARC tasks are fundamentally vision-centric problems relying on spatial priors rather than linguistic logic \citep{hu2025arc}.}
\begin{enumerate}
    \item \textbf{Arithmetic:} 168 items (84 numeric, 84 verbal) requiring \textit{rigid algorithmic processing}, solved by $N=60$ participants.
    \item \textbf{Syllogisms:} 32 problems testing \textit{formal deductive logic}, solved by $N=24$ participants.
    \item \textbf{Logic-ALE:} 20 consistency judgment problems in ALE format (also formal logic), solved by $N=84$ participants.
    \item \textbf{Relational Reasoning:} 84 items testing \textit{transitive inference} ($N=310$).
    \item \textbf{Intuitive Reasoning:} 144 vignettes requiring \textit{common-sense causal judgments} rather than formal rules ($N=58$).
\end{enumerate}

\subsection{Models and Inference}

\begin{figure}[t]
  \centering
  \includegraphics[width=\columnwidth]{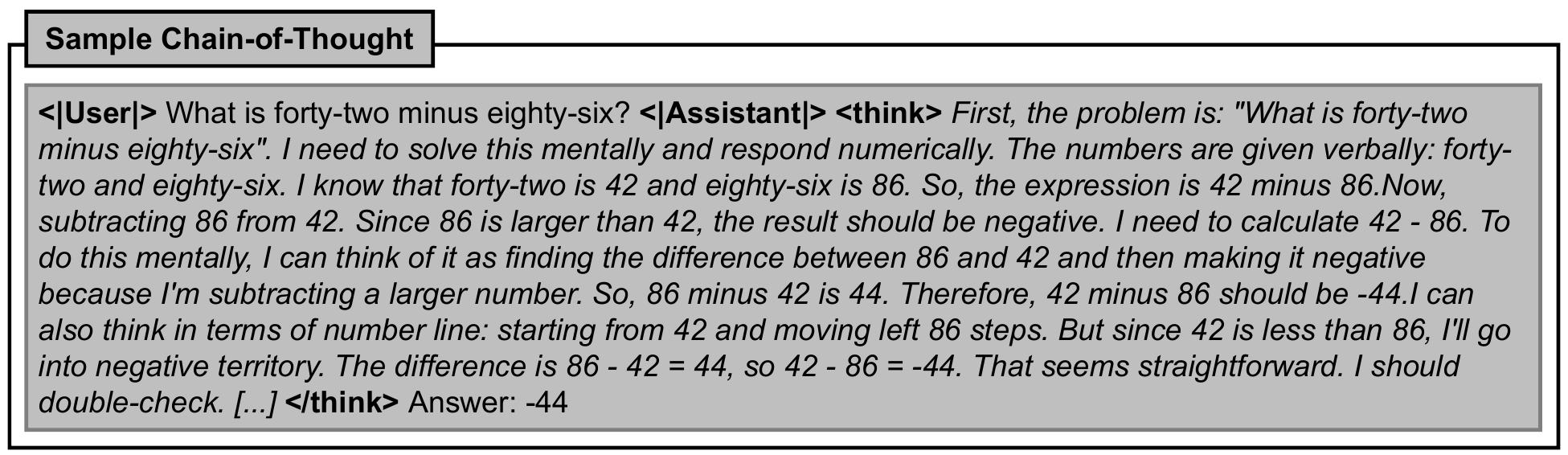}
  \caption{Sample chain-of-thought produced by R1.}
  \label{fig:prompt_strategy}
\end{figure}

To dissect the mechanics of reasoning distillation, we employed a rigorous teacher-student-baseline comparison involving 14 models (see Table~\ref{tab:results_and_models}).

\textbf{Experimental Control: The Role of Instruct Baselines.} A critical confound in evaluating reasoning distillation is distinguishing between the capability to follow instructions (``chat'') and the capability to reason (``think''). Teacher models (e.g., DeepSeek-R1) transmit both linguistic fluency and reasoning structures. To isolate the effect of reasoning distillation, we exclusively selected \textbf{Instruct-tuned} checkpoints as our Base Models (e.g., Qwen2.5-Math-Instruct) rather than raw pre-trained bases. This design ensures that any behavioral divergence observed in the Distilled models is attributable to the internalization of \textit{Chain-of-Thought patterns}, rather than general instruction-following capabilities.

We evaluated three model groups:
\begin{enumerate}
    \item \textbf{Teacher Models:} DeepSeek-R1 and DeepSeek-V3, serving as the gold standard for cognitive alignment.
    \item \textbf{Distilled Models (Student):} Six models from the DeepSeek-R1-Distill family, utilizing Qwen (1.5B, 7B, 14B, 32B) and Llama (8B, 70B) backbones.
    \item \textbf{Base Models:} The six corresponding Instruct-tuned checkpoints: Qwen2.5-Math-Instruct (1.5B, 7B), Qwen2.5-Instruct (14B, 32B), Meta-Llama-3.1-Instruct (8B), and Llama-3.3-Instruct (70B).
\end{enumerate}

Models were accessed via commercial APIs (Aliyun DashScope for Qwen-series, DeepInfra/ModelScope for Llama-series, Together AI for DeepSeek) with temperature $T=0$ to ensure deterministic reproducibility.

\begin{table*}[t]
\centering
\caption{Accuracy (\%) comparisons across six reasoning tasks. Human baselines are adapted from \citet{devarda2025cost}. Models were evaluated using \textbf{Pass@1} with greedy decoding ($T=0$). \textbf{Bold} indicates the best performance in each column; \underline{underline} indicates the best performing Distilled (Student) model.}
\label{tab:results_and_models}
\resizebox{\textwidth}{!}{%
\begin{tabular}{l cc ccccc}
\toprule
 & \multicolumn{2}{c}{\textbf{Arithmetic}} & & & & & \\
\cmidrule(lr){2-3} 
\textbf{Model} & \textit{Numeric} & \textit{Verbal} & \textbf{Syllogism} & \textbf{Logic} & \textbf{Relational} & \textbf{Intuitive} & \textbf{Average} \\
\midrule
Human Baseline & 96.0 & 95.1 & 84.6 & 83.9 & 72.1 & 81.6 & 85.5 \\
\midrule
\multicolumn{8}{l}{\textit{Teacher Models}} \\
DeepSeek-R1 & \textbf{100.0} & 94.2 & \textbf{100.0} & \textbf{100.0} & 94.2 & \textbf{88.9} & \textbf{97.2} \\
DeepSeek-V3 & \textbf{100.0} & \textbf{100.0} & \textbf{100.0} & \textbf{100.0} & 94.2 & 81.2 & 95.9 \\
\midrule
\multicolumn{8}{l}{\textit{Distilled Models (Student)}} \\
DS-R1-Distill (Qwen-1.5B) & 96.4 & 97.6 & 81.2 & 80.0 & 64.0 & 47.2 & 77.7 \\
DS-R1-Distill (Qwen-7B) & 97.6 & 95.2 & \underline{100.0} & \underline{100.0} & 89.5 & 52.8 & 89.2 \\
DS-R1-Distill (Qwen-14B) & \underline{100.0} & \underline{100.0} & \underline{100.0} & 95.0 & \textbf{97.7} & 75.0 & 94.6 \\
DS-R1-Distill (Qwen-32B) & \underline{100.0} & \underline{100.0} & \underline{100.0} & \underline{100.0} & 95.3 & 77.8 & 95.5 \\
DS-R1-Distill (Llama-8B) & 95.2 & 95.2 & 96.9 & 85.0 & 82.6 & 64.6 & 86.6 \\
DS-R1-Distill (Llama-70B) & \underline{100.0} & 98.8 & \underline{100.0} & \underline{100.0} & 94.2 & \underline{85.4} & \underline{96.4} \\
\midrule
\multicolumn{8}{l}{\textit{Base Models (Pre-distillation)}} \\
Qwen2.5-Math (1.5B) & 91.7 & 96.4 & 50.0 & 85.0 & 60.5 & 31.9 & 69.3 \\
Qwen2.5-Math (7B) & 100.0 & 98.8 & 84.4 & 85.0 & 66.3 & 53.5 & 81.3 \\
Qwen2.5 (14B) & 100.0 & 100.0 & 93.8 & 100.0 & 86.0 & 73.6 & 92.2 \\
Qwen2.5 (32B) & 100.0 & 100.0 & 96.9 & 100.0 & 86.0 & 81.2 & 94.0 \\
Llama-3.1 (8B) & 95.2 & 89.3 & 81.2 & 80.0 & 79.1 & 66.7 & 81.9 \\
Llama-3.3 (70B) & 100.0 & 98.8 & 96.9 & 100.0 & 94.2 & 81.9 & 95.3 \\
\midrule
\textit{Distilled Avg} ($\pm$SD) & 98.2 \scriptsize{$\pm$2.1} & 97.8 \scriptsize{$\pm$2.2} & 96.4 \scriptsize{$\pm$7.5} & 93.3 \scriptsize{$\pm$8.8} & 87.2 \scriptsize{$\pm$12.6} & 67.1 \scriptsize{$\pm$15.0} & 90.0 \scriptsize{$\pm$11.9} \\
\textit{Base Avg} ($\pm$SD) & 97.8 \scriptsize{$\pm$3.6} & 97.2 \scriptsize{$\pm$4.1} & 83.9 \scriptsize{$\pm$17.8} & 91.7 \scriptsize{$\pm$9.3} & 78.7 \scriptsize{$\pm$12.9} & 64.8 \scriptsize{$\pm$19.3} & 85.7 \scriptsize{$\pm$12.7} \\
\bottomrule
\end{tabular}%
}
\end{table*}

\subsection{Measures}
For each problem $i$ and model $m$, we extracted:
\begin{itemize}
    \item \textbf{Accuracy ($Acc_{m,i}$):} Validated against ground truth.
    \item \textbf{Reasoning Cost ($C_{m,i}$):} For R1 and distilled models, cost is defined as the token count within the \texttt{<think>} delimiters (see Figure~\ref{fig:prompt_strategy}). For base models, we appended the CoT trigger ``\textit{Let's think step by step}'' and counted the full completion length.
\end{itemize}

\subsection{Analysis Framework}
We structured our analysis to distinguish between superficial imitation (\textit{Mimicry}) and functional internalization (\textit{Mastery}).

\subsubsection{1. Functional Alignment}
We assessed whether models retain human-like difficulty scaling by computing the Pearson correlation ($r_{align}$) between the log-transformed reasoning cost and human reaction times ($RT$):
\begin{equation}
    r_{align} = \text{Corr}(\log(C_{m}), \log(RT_{human}))
\end{equation}

\subsubsection{2. Representational Similarity Analysis (RSA)}
To quantifiably map the topology of reasoning strategies, we constructed ``difficulty fingerprints'' for all agents. We first log-transformed reasoning costs and z-scored them \textit{within each task} to isolate relative difficulty patterns. Formally, the fingerprint $\mathbf{v}_m$ for model $m$ is the concatenation of standardized costs across tasks:
\begin{equation}
    \mathbf{v}_m = \bigoplus_{t \in T} \mathcal{Z}(\log(\mathbf{c}_{m,t}))
\end{equation}
We utilized these fingerprints to compute pairwise Pearson correlations, generating the RSA matrix. Furthermore, to visualize the distillation geometry, we projected models into a 2D space defined by \textbf{Teacher Similarity} ($x = \text{Corr}(\mathbf{v}_S, \mathbf{v}_{Teacher})$) and \textbf{Human Alignment} ($y = \text{Corr}(\mathbf{v}_S, \mathbf{v}_{Human})$).

\subsubsection{3. Surface Similarity Metrics}
To quantify how closely students mimic the teacher's output distribution while accounting for varying task lengths, we first calculated the \textbf{Relative Effort} ($e_{m,i}$) by normalizing the token cost of each item by the teacher's average cost for that task ($e_{m,i} = C_{m,i} / \bar{C}_{T,\text{task}}$). We then computed two metrics over the distributions of $e$:

\begin{equation}
    \rho = \mu(e_{S}), \quad D_{KL}(P || Q) = \sum_{x} P(x) \log \left( \frac{P(x)}{Q(x)} \right)
\end{equation}

where $\rho$ denotes the \textbf{Thinking Effort Ratio} (measuring verbosity inflation), and $D_{KL}$ denotes the \textbf{Strategy Divergence} between the probability distribution of the Teacher's effort ($P$) and the Student's effort ($Q$).

\subsubsection{4. Structural Mechanism: The Linear Inflation Law}
To determine the generative mechanism of efficiency degradation, we first formulated the \textbf{Inverse Efficiency Index} ($\mathcal{I}_{E}$) as the ratio of average reasoning cost ($\bar{C}$) to accuracy ($Acc$):
\begin{equation}
    \mathcal{I}_{E} = \frac{\bar{C}_{m}}{Acc_{m}}
\end{equation}
We then tested for a \textbf{proportional Linear Inflation Law} by modeling the relationship between student and base models via a regression constrained to the origin:
\begin{equation}
    \mathcal{I}_{E}^{student} \approx N \cdot \mathcal{I}_{E}^{base}
\end{equation}
where the slope $N$ represents a ``verbosity multiplier.'' A strict fit to this model implies that distillation imposes a pure multiplicative penalty on computational cost, independent of the base model's intrinsic heuristics.

\section{Results}

\subsection{The Phenomenology of Failure: Diagnosis of ``Hán Dān Xué Bù''}

\begin{figure}[t!]
  \centering
  \includegraphics[width=\linewidth]{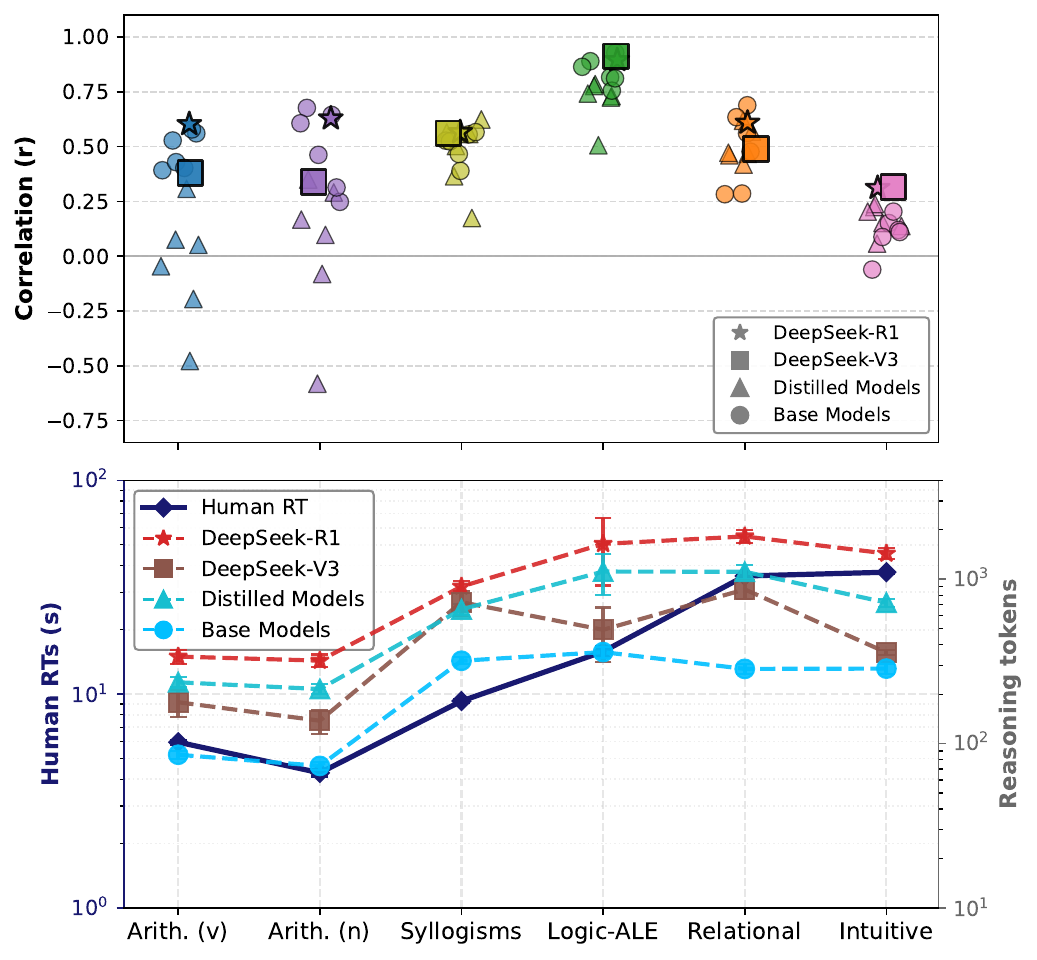}
  \caption{\textbf{Combined Analysis of Cognitive Alignment and Reasoning Cost.} 
(Top) Correlation ($r$): between reasoning cost and human RTs. (Bottom) Comparison of Human RTs (solid, left axis) and model token counts (dashed, right axis) across tasks.}
  \label{fig:combined_results}
\end{figure}

\paragraph{Functional Alignment Collapse}
Reasoning distillation induces a severe collapse in cognitive alignment. While the Teacher model (DeepSeek-R1) mirrors human difficulty scaling ($\bar{r} = 0.640$), Distilled models significantly degrade this alignment ($\bar{r} = 0.341$, $t=2.45, p=0.044$).

Crucially, we observe a \textbf{``Negative Transfer''} effect confirmed by statistical testing ($t=-2.60, p=0.011$). Distilled models perform significantly \textit{worse} than their own Base models ($\bar{r}_{base} = 0.534$) and fail to match the baseline alignment of standard RLHF models like DeepSeek-V3 ($\bar{r}=0.558$). This indicates that the distillation process effectively ``overwrites'' the base models' native, efficient heuristics with a maladaptive strategy (see Figure~\ref{fig:combined_results}, top panel), supporting the premise that the student has lost its original ``way of walking.''

\begin{figure}[t!]
  \centering
  \includegraphics[width=\columnwidth]{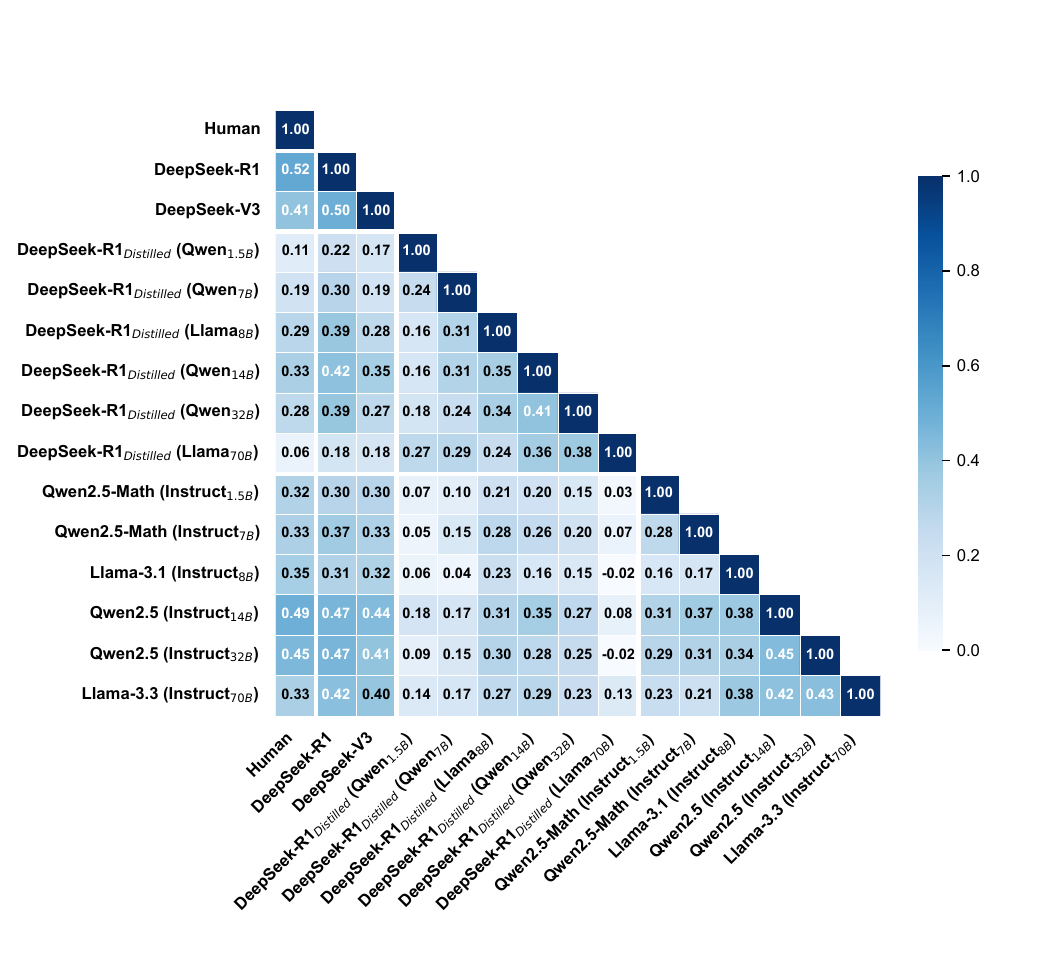}
  \caption{\textbf{Representational Similarity Analysis (RSA) Heatmap.} Values denote Pearson correlations of difficulty fingerprints (z-scored costs). Distilled models exhibit low structural similarity to the Teacher (DeepSeek-R1), indicating a failure to internalize the target reasoning policy.}
  \label{fig:rsa_heatmap}
\end{figure}

\paragraph{Topological Drift: The ``Spurious Valley''}
To map the structural consequences of this collapse, we employed Representational Similarity Analysis (RSA). RSA reveals a stark representational drift (Figure~\ref{fig:rsa_heatmap}): while Human and Teacher patterns are structurally similar ($r=0.52$), distilled models become untethered from both.

\begin{figure}[t!]
  \centering
  \includegraphics[width=\columnwidth]{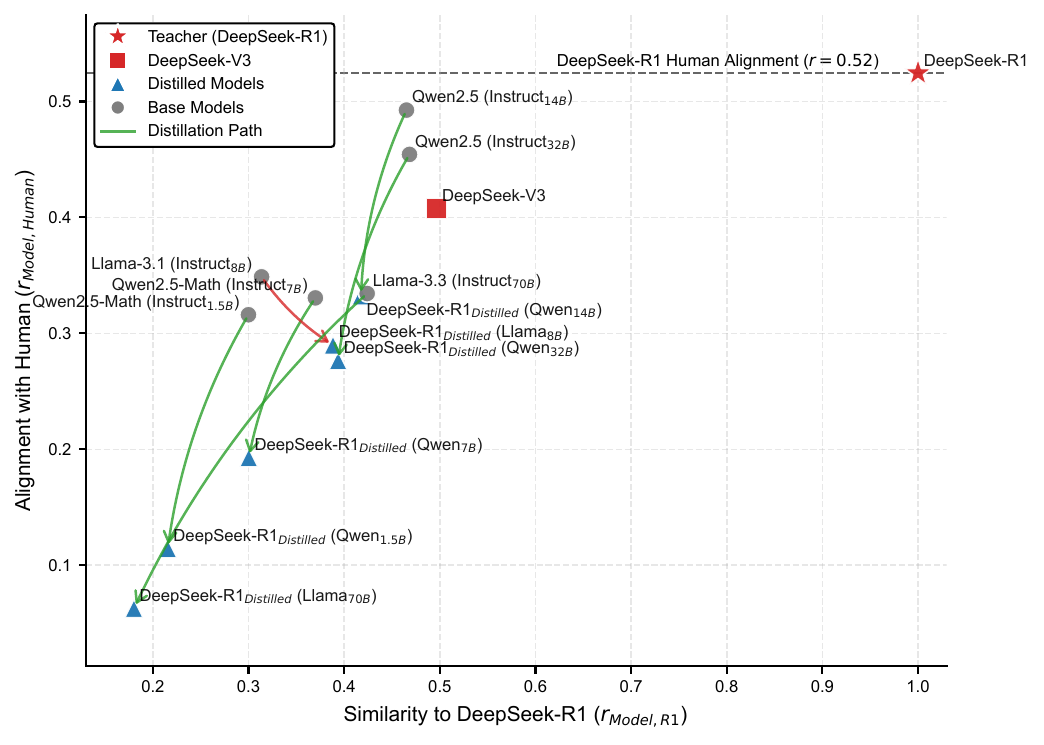}
  \caption{\textbf{The Cognitive Waterfall.} Models define a topology of failed imitation defined by Teacher Similarity ($x$-axis) and Human Alignment ($y$-axis). Instead of converging to the Teacher (top-right), distilled models collapse into a ``Spurious Valley'' characterized by simultaneous loss of alignment (The Drop) and structural similarity (The Drift).}
  \label{fig:triangle_scatter}
\end{figure}

Figure~\ref{fig:triangle_scatter} illustrates the \textbf{``Cognitive Waterfall.''} Instead of converging to the Teacher, student models collapse into a ``Spurious Valley'' driven by two vectors:

\begin{enumerate}
    \item \textbf{The Drop (Alignment Collapse):} A vertical descent indicating a loss of human-likeness. In terms of representational similarity to humans, distilled models score significantly lower ($\bar{r} = 0.21$) than their base counterparts ($\bar{r} = 0.38$, $p = 0.01$).
    
    \item \textbf{The Drift (Similarity Regression):} A counter-intuitive leftward shift indicating that distillation actively \textit{degrades} structural similarity to the teacher. Quantitatively, distilled models show lower representational correlation with the Teacher ($\bar{r} = 0.32$) than even their raw, pre-distilled base counterparts ($\bar{r} = 0.39$), showing a marginally significant trend ($p = 0.09$).
\end{enumerate}

Together, these results confirm that distilled models occupy a ``Spurious Valley''---a cognitive state distinct from both human intuition and robust machine reasoning.

\begin{figure}[t!]
  \centering
  \includegraphics[width=\columnwidth]{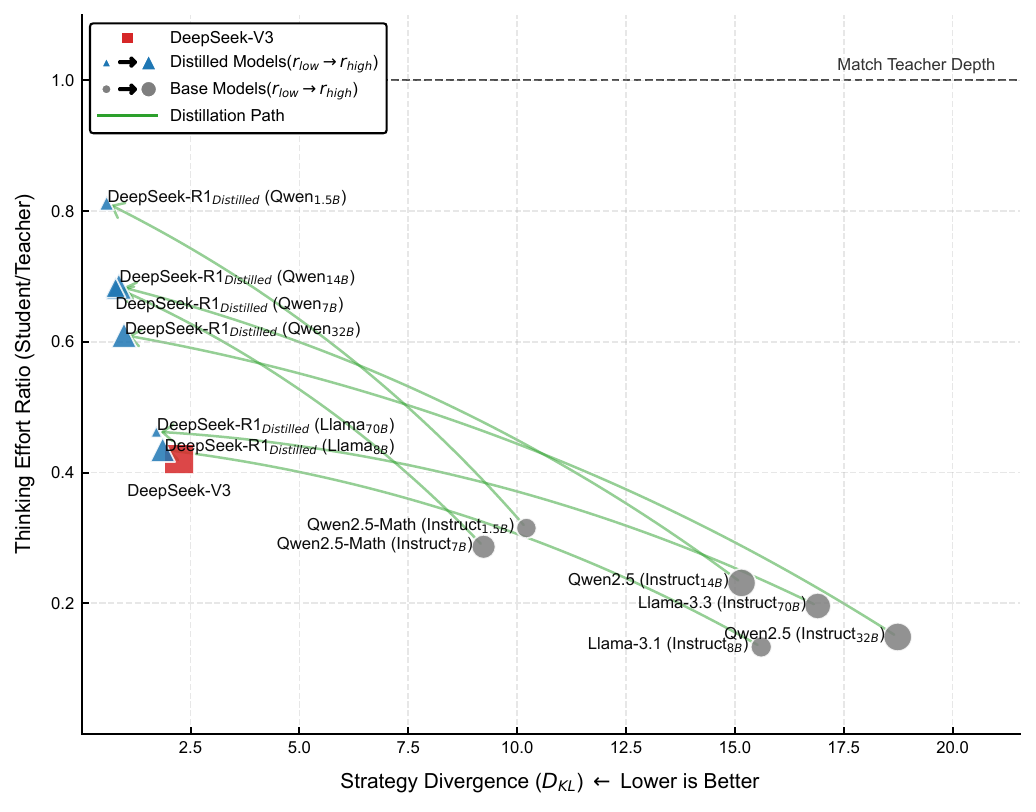}
  
  \vspace{0.2cm} 
  
  \includegraphics[width=\columnwidth]{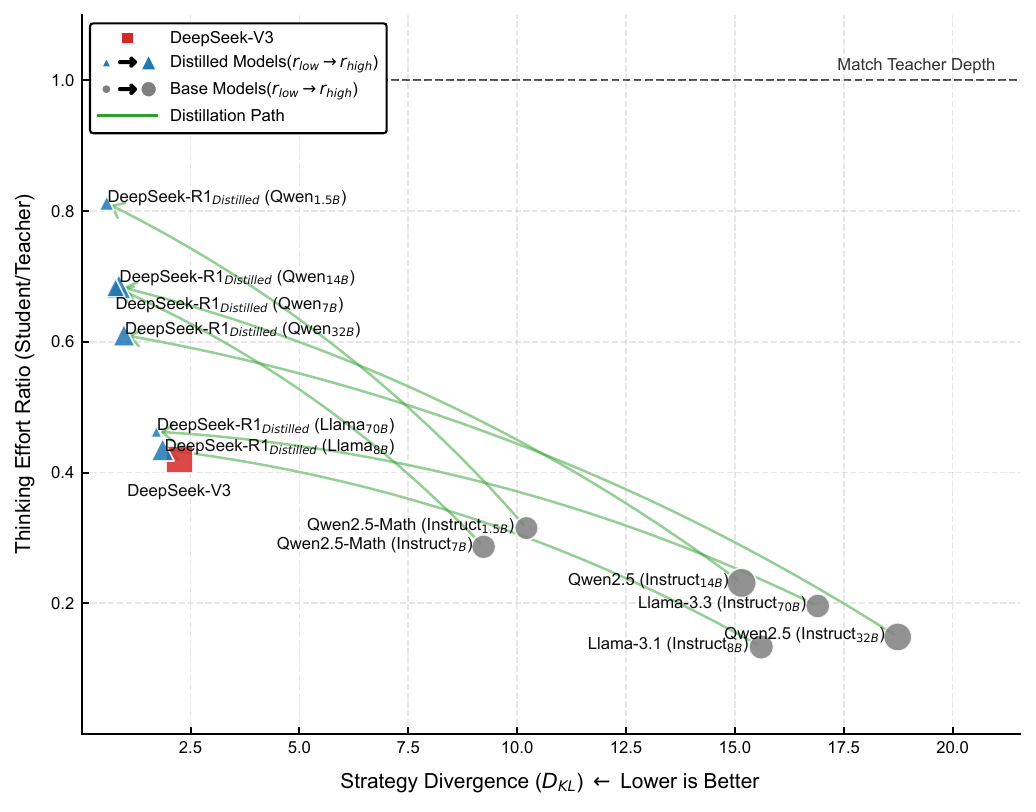}
  
  \caption{\textbf{Surface-Level Mimicry Landscape.} Models migrate to the high-verbosity ``Mimicry Region'' (top-left, low $D_{KL}$ and $\rho \approx 1.0$). Bubble sizes indicate (a, top) representational similarity to the Teacher and (b, bottom) alignment with Human reaction times. Shrinking bubbles in \textit{both} panels reveal that surface-level mimicry decouples computational effort from both the Teacher's item-wise allocation policy and human cognitive demand.}
  \label{fig:landscape_combined}
\end{figure}

\subsection{The Mechanism of Mimicry: Quantitative Analysis}

To investigate the generative mechanisms of the collapse, we first analyzed surface-level metrics.
\paragraph{Surface-Level Divergence.} 
We computed the Thinking Effort Ratio ($\rho$) and Strategy Divergence ($D_{KL}$) to assess distributional mimicry. As illustrated in Figure~\ref{fig:landscape_combined}, distillation effectively minimizes the Kullback-Leibler divergence, driving student models into a ``Mimicry Region'' characterized by low $D_{KL}$ and a Teacher-like effort density ($\rho \approx 1.0$). 

However, visual inspection of Figure~\ref{fig:landscape_combined} reveals a critical discrepancy. In panel (a), bubbles shrink as students enter the Mimicry Region, indicating that despite matching the Teacher's \textit{aggregate} effort distribution (low $D_{KL}$), distilled students diverge from the Teacher's \textit{item-wise} difficulty fingerprint ($\bar{r}_{Model,R1} = 0.32$; cf.\ Figure~\ref{fig:rsa_heatmap}). In panel (b), bubbles shrink further, showing that alignment with human reaction times degrades even more ($\bar{r}_{Model,Human} = 0.21$). This \textit{dual decoupling}---from both the Teacher's allocation policy and human cognitive demand---demonstrates that distillation transmits the statistical form of reasoning traces without their underlying cognitive function.

\begin{figure*}[t!]
  \centering
  \includegraphics[width=\textwidth]{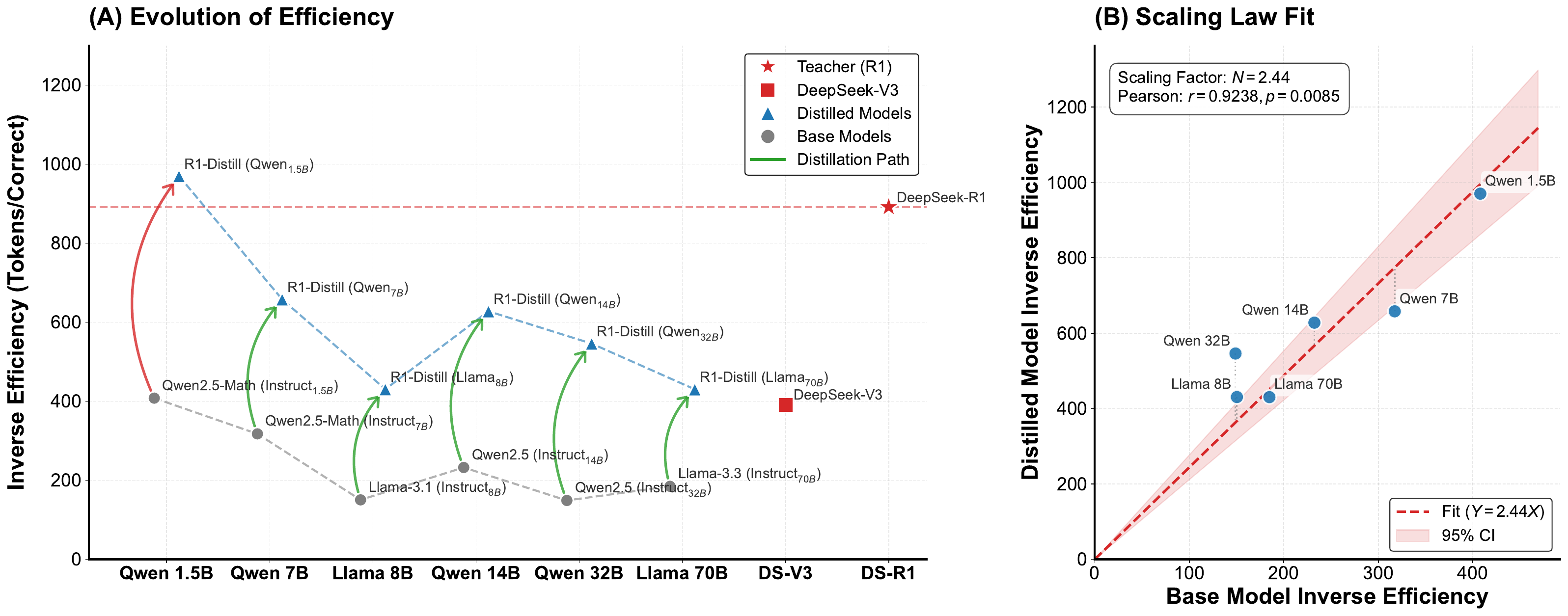}
  \caption{\textbf{Inverse Efficiency Analysis.} \textbf{(A)} Distilled models (blue triangles) exhibit consistently worse efficiency (tokens/correct) than their base counterparts (grey circles). \textbf{(B)} Regression confirms a Linear Inflation Law: distillation imposes a systematic multiplicative penalty ($N \approx 2.44$) on the base model's inefficiency, independent of intrinsic reasoning heuristics.} 
  \label{fig:efficiency}
\end{figure*}

\paragraph{The Linear Inflation Law.}
We further investigated intrinsic efficiency changes by modeling the Inverse Efficiency Index ($\mathcal{I}_{E}$). Regression analysis (Figure~\ref{fig:efficiency}) reveals a strict linear dependency between the inefficiencies of student and base models ($Pearson's~r = 0.92, p < 0.01$). We confirm the \textbf{Linear Inflation Law} proposed in Eq.~(5), finding a fitted slope of $N \approx 2.44$. This result suggests that distillation imposes a systematic, multiplicative penalty on computational cost. Rather than adopting the Teacher's dynamic allocation policy, student models mechanically inflate their base output length by a fixed factor of $\sim2.44$ across tasks, regardless of problem complexity.

\section{Discussion}

Our findings reveal a paradox: while student models statistically reproduce the teacher's verbose traces, they fail to internalize the underlying cognitive policy driving that verbosity. This disconnect—the \textit{Hán Dān Xué Bù} effect—exposes a fundamental epistemic limit in current reasoning distillation paradigms.

\subsection{The Illusion of Depth: Why Distillation Degrades Alignment}
A critical anomaly in our results is the ``Negative Transfer'' effect: distilled models exhibit significantly worse cognitive alignment ($\bar{r}=0.34$) than their own instruction-tuned base models ($\bar{r}=0.53$). Since both groups utilize Supervised Fine-Tuning (SFT), the failure is not inherent to SFT but specifically arises from \textit{CoT Distillation}.

Standard instruction tuning (Base Models) typically optimizes for answer correctness across diverse tasks, allowing the model to utilize its native, albeit simpler, heuristics. In contrast, CoT distillation forces the student to mimic the teacher's extended ``thinking process'' as a target output. This triggers a form of catastrophic interference \citep{french1999catastrophic}, where verbose patterns overwrite the base model's efficient latent structures. Unlike biological complementary learning systems \citep{mcclelland1995there} that integrate new knowledge without erasing the old, the distilled models succumb to ``Negative Transfer,'' performing worse than their pre-distillation baselines.

Furthermore, our ``Linear Inflation Law'' ($N \approx 2.44$) suggests that students treat reasoning tokens not as functional steps, but as a rigid stylistic constraint. This mirrors the Einstellung (set) effect \citep{bilalic2008good, bilalic2010mechanism}: the student becomes ``blinded'' by a familiar, complex strategy (verbosity), mechanically applying it even when simpler heuristics would suffice. The result is a ``fluency illusion'' \citep{deslauriers2019measuring}: the model's coherent output masks a lack of cognitive engagement, creating a false appearance of competence.

\subsection{The Credit Assignment Problem in Passive Learning}
Why does mimicking the teacher's process lead to functional collapse? We propose that reasoning emerges from \textit{credit assignment}—a mechanism SFT structurally lacks. In the teacher's Reinforcement Learning (RL) phase, verbosity is a byproduct of a search process surviving sparse rewards; a long chain exists only because it was instrumental in reaching a correct solution. The teacher effectively performs a neuroeconomic calculation, scaling cognitive effort where the expected value of correctness outweighs the computational cost \citep{westbrook2015cognitive}.

In SFT, however, every token is weighted equally; students receive identical updates for both vacuous fillers and crucial logical pivots. Consequently, the student fails to develop critical meta-reasoning capabilities—specifically the monitoring and control of cognitive effort \citep{ackerman2017meta}. Lacking internal error signals or the feeling of rightness'' that guides human termination of thought, the student cannot distinguish between the scaffolding'' of thought and the thought itself. This aligns with the ``Spring Model'' of alignment \citep{ji2025resist}: distillation stretches the model's output distribution to match the teacher's length (elasticity) but fails to restructure its internal plasticity, resulting in the ``Spurious Valley'' observed in our RSA analysis. This account dovetails with an ongoing debate. \citet{vankov2026correlations} and \citet{hu2026thinking} argued that LRM trace length may reflect \textit{performative scaffolding} rather than genuine computation, while \citet{devarda2026replyvankov} showed that RLVR traces track validated difficulty dimensions. Our findings dissociate these positions: the alignment signal survives under RLVR but collapses under CoT distillation, with the Linear Inflation Law as the quantitative signature of the scaffolding regime.

\subsection{From Behavioral Mimicry to Cognitive Mastery}
The contrast between our RL-trained teacher and SFT-trained students parallels, \textit{heuristically}, the distinction between \textit{behaviorist} conditioning and \textit{constructivist} learning---with the caveat, following \citet{brady2025dual}, that such framings serve to interpret outputs rather than assert shared cognitive architecture. SFT assumes that reasoning can be transmitted via passive observation---watching the teacher walk. However, cognitive skill acquisition theories posit that declarative knowledge (traces) does not convert to procedural knowledge (reasoning) without active practice \citep{anderson1982acquisition, vanlehn1996cognitive}. Our data suggests that ``System 2'' reasoning \citep{kahneman2011thinking} is not a static template to be memorized, but a dynamic policy for resource allocation that requires the proceduralization of heuristics.

True mastery (\textit{Q\=ing Ch\=u Y\'u L\'an}) likely requires the agent to engage in ``cognitive foraging''---actively exploring the search space and experiencing the error signals necessary to prune inefficient paths. This aligns with findings that self-directed learning facilitates structural generalization significantly better than passive reception \citep{gureckis2012self}. While SFT provides the necessary linguistic scaffolding, akin to Vygotsky's Zone of Proximal Development \citep{vygotsky1978mind}, transcending the teacher's capabilities requires the autonomous pressure of rewards. 

Future cognitive modeling must therefore transcend ``trace cloning,'' utilizing methods that incentivize students to rediscover the \textit{function} of reasoning. Specifically, we propose training on traces only when they yield verified correct answers, enabling the model to experience its own success, or employing teacher traces as a \textit{prior} within reinforcement learning rather than as static scripts.

\section{Acknowledgments}

We thank Dr.\ Huzi Cheng, Jiaxin Li, and Junxian Yang for insightful discussions on the use of Reinforcement Learning (RL) and Supervised Fine-Tuning (SFT) techniques in training Large Language Models (LLMs), which substantially informed the framing of this work. We acknowledge the use of Gemini 3 Pro to assist with the development of experimental scripts, data analysis code, and visualization pipelines. Additionally, the model was utilized for linguistic refinement and copy-editing of the manuscript. The authors retain full responsibility for the scientific content and final text.





\printbibliography

\end{document}